\documentclass[]{interact}

\usepackage{epstopdf}
\usepackage[caption=false]{subfig}

\usepackage{natbib}
\bibpunct[, ]{(}{)}{;}{a}{}{,}
\renewcommand\bibfont{\fontsize{10}{12}\selectfont}

\theoremstyle{plain}

\theoremstyle{definition}

\theoremstyle{remark}

\begin{document}


\title{3D Human Pose Estimation in Multi-View Operating Room Videos Using Differentiable Camera Projections}

\author{
    \name{Beerend~G.A.~Gerats\textsuperscript{a,b}\thanks{CONTACT B.G.A. Gerats. Email: bga.gerats@meandermc.nl}, Jelmer~M.~Wolterink\textsuperscript{b} and Ivo~A.M.J.~Broeders\textsuperscript{a,b}}
    \affil{
        \textsuperscript{a}Centre for Artificial Intelligence, Meander Medisch Centrum, Amersfoort, the Netherlands \textsuperscript{b}University of Twente, Enschede, the Netherlands
    }
}

\maketitle

\begin{abstract}
3D human pose estimation in multi-view operating room (OR) videos is a relevant asset for person tracking and action recognition. However, the surgical environment makes it challenging to find poses due to sterile clothing, frequent occlusions, and limited public data. Methods specifically designed for the OR are generally based on the fusion of detected poses in multiple camera views. Typically, a 2D pose estimator such as a convolutional neural network (CNN) detects joint locations. Then, the detected joint locations are projected to 3D and fused over all camera views. However, accurate detection in 2D does not guarantee accurate localisation in 3D space. In this work, we propose to directly optimise for localisation in 3D by training 2D CNNs end-to-end based on a 3D loss that is backpropagated through each camera's projection parameters. Using videos from the MVOR dataset, we show that this end-to-end approach outperforms optimisation in 2D space.
\end{abstract}


\section{Introduction}

\begin{figure}
    \includegraphics[width=\textwidth]{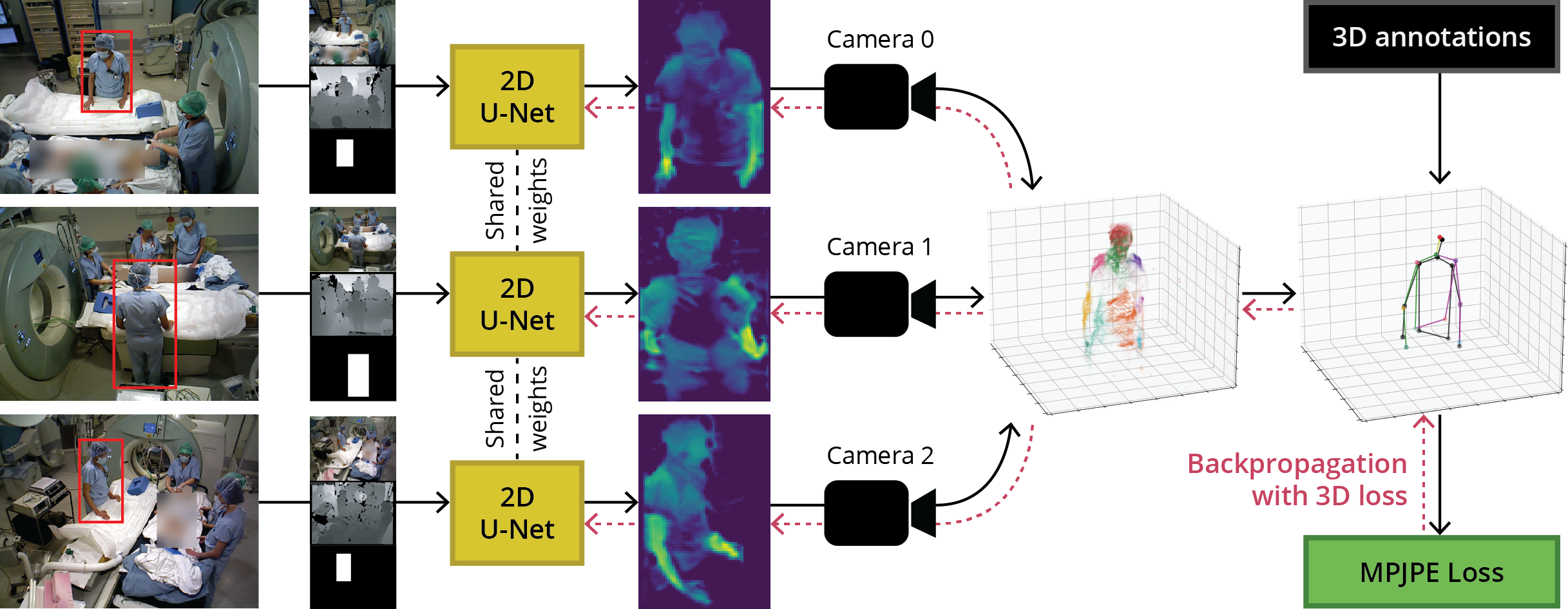}
    \caption{Overview of the proposed method. Up to three 2D input images are each processed by a single U-Net that generates pixel-wise predictions for the presence of body joints. All pixels are projected to a shared 3D coordinate system, a centre of mass is computed for each body joint in 3D, and a loss is determined by comparison to a reference pose via a 3D distance metric (MPJPE). The 2D U-Nets are directly optimised using this 3D loss by backpropagation through the camera projections.} \label{fig:overview}
\end{figure}

Automated estimation of human poses in 3D in multi-view operating room (OR) videos is a relevant asset in surgical data science \citep{maier2022surgical}. Information about people's poses could help to explain surgical phases~\citep{schmidt2021} and actions of medical staff~\citep{twinanda2015}. In addition, pose analysis could form the basis for workflow recognition of the surgical team and for X-ray radiation monitoring to enhance staff safety~\citep{padoy2019}. With the advent of deep learning, the last decade has seen significant improvements in human pose estimation (HPE), e.g. of fast-moving people in busy crowds~\citep{cao2019,fang2017}. However, the surgical environment imposes specific challenges that limit the applicability of off-the-shelf pose estimators, requiring methods that are explicitly designed for the OR.

Challenges that apply to OR videos include people wearing sterile clothing, facial masks, hairnets, and eye protection, which drastically differs from ordinary clothing. Moreover, instruments such as surgical lamps, hanging screens, and monitoring devices cause occlusions that regularly make people invisible to the camera. These instruments do not have a fixed position and can be moved throughout a surgical procedure. Lastly, the sensitive character of surgical videos makes it challenging to acquire large annotated datasets. The limited amount of publicly available data is likely to bound the performance of deep learning methods, which benefit from large datasets~\citep{sun2017}. To overcome these challenges, clinical pose estimation requires methods that can be trained with only small amounts of data.

Existing works on 3D HPE in OR videos typically use synchronised colour or depth video streams from multiple camera angles \citep{belagiannis2016}, such as those in the public \textit{Multi-View Operating Room} (MVOR) dataset~\citep{srivastav2018}. Generally, two-step approaches are used. First, a 2D pose estimator detects body joint locations or generates probability heatmaps for individual views~\citep{belagiannis2016,kadkhodamohammadi2017}. Second, the locations are projected to 3D, fused overall views, and post-processed~\citep{hansen2019,kadkhodamohammadi2021}. However, localisation in individual 2D views might not necessarily lead to optimal localisation in 3D.

We propose a novel approach to pose estimation in multi-view OR videos. We train 2D U-Nets for joint localisation in camera views, but use differentiable camera projections and pose fusion in 3D to directly compute a loss with respect to the 3D reference pose. Because this forward pass is differentiable, we can train the whole model end-to-end, i.e., we indirectly optimise the parameters of the 2D U-Nets based on a 3D loss (see Fig.~\ref{fig:overview}). We show that this approach improves 3D joint localisation for more accurate pose estimation. Our contributions are: (1) the presentation of a novel approach to 3D HPE in multi-view OR videos that can be trained end-to-end based on a 3D loss, (2) an approach for invertible data augmentation in multi-view datasets, and (3) an empirical comparison between the use of colour and depth images as input data modality for pose estimation in the OR.

\section{Methods}
Figure~\ref{fig:overview} shows an overview of the proposed, fully differentiable model. We consider a multi-view setting, where $V$ cameras simultaneously capture the same scene from different angles. In this scene, there are $N$ persons present. Each person is defined as a pose, consisting of $J$ body joints, for which ground-truth locations are available in three dimensions: $p_j^n \in \mathbb{R}^3$ for all $j \in \{1, 2, \ldots, J\}$ and $n \in \{1, 2, \ldots, N\}$. We aim to estimate the locations of all body joints such that the average 3D Euclidean distance between the estimated joint locations $\hat{p}_j^n$ and annotated joint locations $p_j^n$ is minimised. This distance metric is generally known as the \textit{Mean Per Joint Prediction Error} (MPJPE)~\citep{li2015} and is calculated as in Equation~\ref{eq:mpjpe}. The design of a model architecture that outputs 3D body joint locations gives the advantage that we can use MPJPE as both the loss function and the evaluation metric.

\begin{equation}
    \text{MPJPE} = \dfrac{1}{JN} \sum_{j=1}^J \sum_{n=1}^N \left\Vert \hat{p}_j^n - p_j^n \right\Vert_2.
    \label{eq:mpjpe}
\end{equation}

We train a single U-Net \citep{ronneberger2015} per camera view to generate a localisation heatmap for each body joint in a 2D view. The U-Nets have shared weights and are optimised simultaneously. As input, each U-Net takes a three-channel RGB image, a one-channel depth image, and a one-channel binary image indicating the location of a bounding box around a person. These are stacked to obtain an input tensor with five channels. The final layer of the U-Net is linear such that the output values in the localisation heatmap correspond to the proximity to a particular body joint. This means that a U-Net has $J$ output channels, one for each body joint type.

\subsection{Differentiable Camera Projections}
Although the heatmaps provided by the U-Nets indicate the presence of a body joint in an area, they are only estimates in 2D. To find the positions in 3D, all pixel activation values from the U-Net output are projected from 2D to a 3D coordinate system that is shared for all views. For each person in the scene, we do the following. First, in each view, we set pixel activations outside that person's bounding box to a negative value $\epsilon$. Similarly, we set pixel activations for which the depth is unknown to $\epsilon$. Second, all pixel activation values are projected from 2D to 3D along the camera rays according to the corresponding depth image $\mathcal{D}^v$ using the intrinsic parameters (i.e. principal point and focal length) of each camera. This transformation can be expressed as:

\[
    (x', y', z')^v = \Big( \mathcal{D}^v(x, y)\dfrac{x - r_x^v}{f_x^v}, \mathcal{D}^v(x, y)\dfrac{y - r_y^v}{f_y^v}, \mathcal{D}^v(x, y) \Big),
\]

with $(x', y', z')^v \in \mathbb{R}^3$ the projected 3D location in coordinate system of view $v \in \{1, 2, \ldots, V\}$, $(x, y)$ the 2D location of a pixel in the original image, $\mathcal{D}^v(x,y)$ the depth value of that pixel in the depth image, $r^v$ and $f^v$ the principal point and focal length of camera $v$. Third, we project all points to the coordinate system of the first camera, based on the extrinsic camera parameters, by:

\[
    \begin{bmatrix} \hat{x} & \hat{y} & \hat{z} & 1 \end{bmatrix} = T^{v\rightarrow 0} \begin{bmatrix} x' & y' & z' & 1 \end{bmatrix}^T,
\]

with $T^{v \rightarrow 0} \in \mathbb{R}^{4 \times 4}$ the transformation matrix from camera $v$ to camera 0 and $(\hat{x}, \hat{y}, \hat{z}) \in \mathbb{R}^3$ the projected locations in the shared coordinate system. Importantly, all operations are based upon matrix multiplication, division, and subtraction, which are differentiable functions. In this way, our method can backpropagate through these operations to optimize the 2D U-Nets whose output is the input for these camera projections.

\subsection{Obtaining 3D Poses}
Projecting all pixels in the 2D heatmaps to 3D results in one 3D point cloud per body joint and view. Point clouds from different views that refer to the same body joint are combined, such that they consist of $I = V \times H \times W$ points, with $H$, $W$ the height and width of the input images. For each body joint, we obtain its estimated 3D location $\hat{p}_j^n$ as a weighted sum of the coordinates of all points in its combined point cloud. The weighting is defined by the predictions of the U-Nets for each of the points in the point cloud. We apply a \textit{softmax} function to all points in the point cloud to ensure that their weights $a_i$ for all $i \in \{1, 2, \ldots, I\}$ sum up to 1. Hence, a prediction is obtained as follows:

\[
    \hat{p}_j^n = \sum_{i=1}^I (\hat{x}_i, \hat{y}_i, \hat{z}_i) a_i.
\]

Note that this aggregation procedure is composed of summations and products and is therefore also differentiable. The estimated joint location is used to compute the MPJPE loss (Equation~\ref{eq:mpjpe}). Consequently, this loss is backpropagated through the point cloud aggregation function, through the camera projections and through the layers of the U-Nets.

\subsection{Data Augmentation}
A common approach to reducing the generalisation error of pose estimation via deep learning is to use data augmentation. Data augmentation in the monocular setting is trivial since the same transformations that are applied to the data can be applied to the labels. However, augmentation in our multi-view setting is more challenging. For example, horizontally flipping one or multiple views would break the transformations to the shared coordinate system. This calls for dedicated data augmentation to multi-view datasets with 3D targets.

We apply data augmentation to the input tensor that contains colour images, depths and bounding box masks. The augmentation is applied to each camera separately before the images are given to the U-Nets. Whenever an augmentation is applied to a view, the inverse of the augmentation is applied to U-Net's output. As long as the augmentation and its inverse are matrix multiplications, the whole operation is differentiable such that the model can backpropagate its loss through the inverse augmentation. In our experiments, we use horizontal flipping, in which we simply apply the same matrix multiplication twice, random cropping, which we can invert by zero-padding to the original size, and rotation, where we take the inverted rotation matrix for inverting the augmentation.

\section{Experiments and Results}

\subsection{Data}
We use the MVOR dataset \citep{srivastav2018} to develop and validate our method. This set contains colour and depth image recordings that are synchronously captured from three camera angles in the OR. The authors recorded surgeries in four days and collected 57, 330, 223 and 122 static multi-view images for day one to four, respectively. We sub-sample the images with a factor of 2, such that their resolution is: $240 \times 320$ pixels. The set contains 1061 upper body poses, annotated in 2D and 3D. The poses consist of ten body joints: head, neck, shoulders, hips, elbows and wrists, where the latter four have a left and a right part.

\subsection{Experimental Setup}
We apply four-fold cross-validation, where the images from one day are held out as the test set in each fold. All reported scores are averages of the four folds. We optimise all ten body joints with equal weight. The models are trained for 200 epochs, with a batch size of 1 scene, a horizontal flip probability of 0.5, random cropping to images of $208 \times 288$ pixels, random rotation between -15 and 15 degrees, and an Adam optimiser \citep{kingma2017} with a learning rate of 1e-3. The hyperparameter choices are based upon manual tuning within reasonable boundaries. Parameters were changed in isolation and its effect was assessed on a fixed test set (all samples in day 4).

In the first part of our experiments, our focus is on the \textit{pose estimation} of each person, not their \textit{detection}. Hence, similarly to related work \citep{hansen2019}, reference bounding boxes for persons in camera views are used as one of the input channels to the U-Nets. This allows evaluation independent of the choice for a person detector. In the second part, including Sections~\ref{sec:detr} and \ref{sec:qualitative}, we use automatically detected bounding boxes to explore generalisation to a more realistic scenario in which hand-drawn boxes are not available for new and unseen videos.

\begin{figure}
    \centering
    \subfloat[Projection of 2D skeletons to 3D. For each body joint, the centre of mass is calculated in 2D.]{%
        \resizebox*{10cm}{!}{\includegraphics{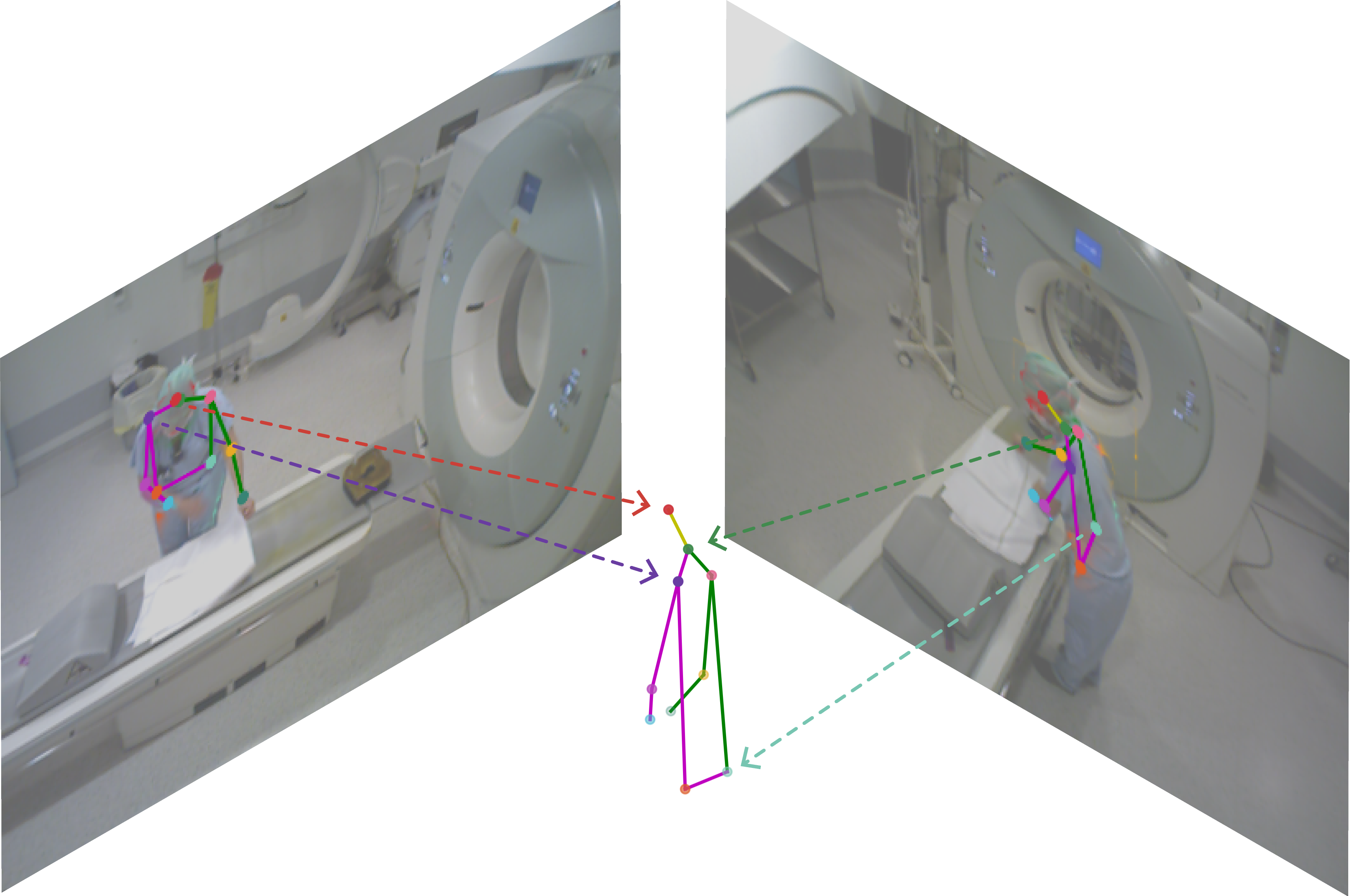}}}
        
    \subfloat[Projection of all heatmap pixels into a point cloud. For each body joint, the centre of mass is calculated in 3D.]{%
        \resizebox*{10cm}{!}{\includegraphics{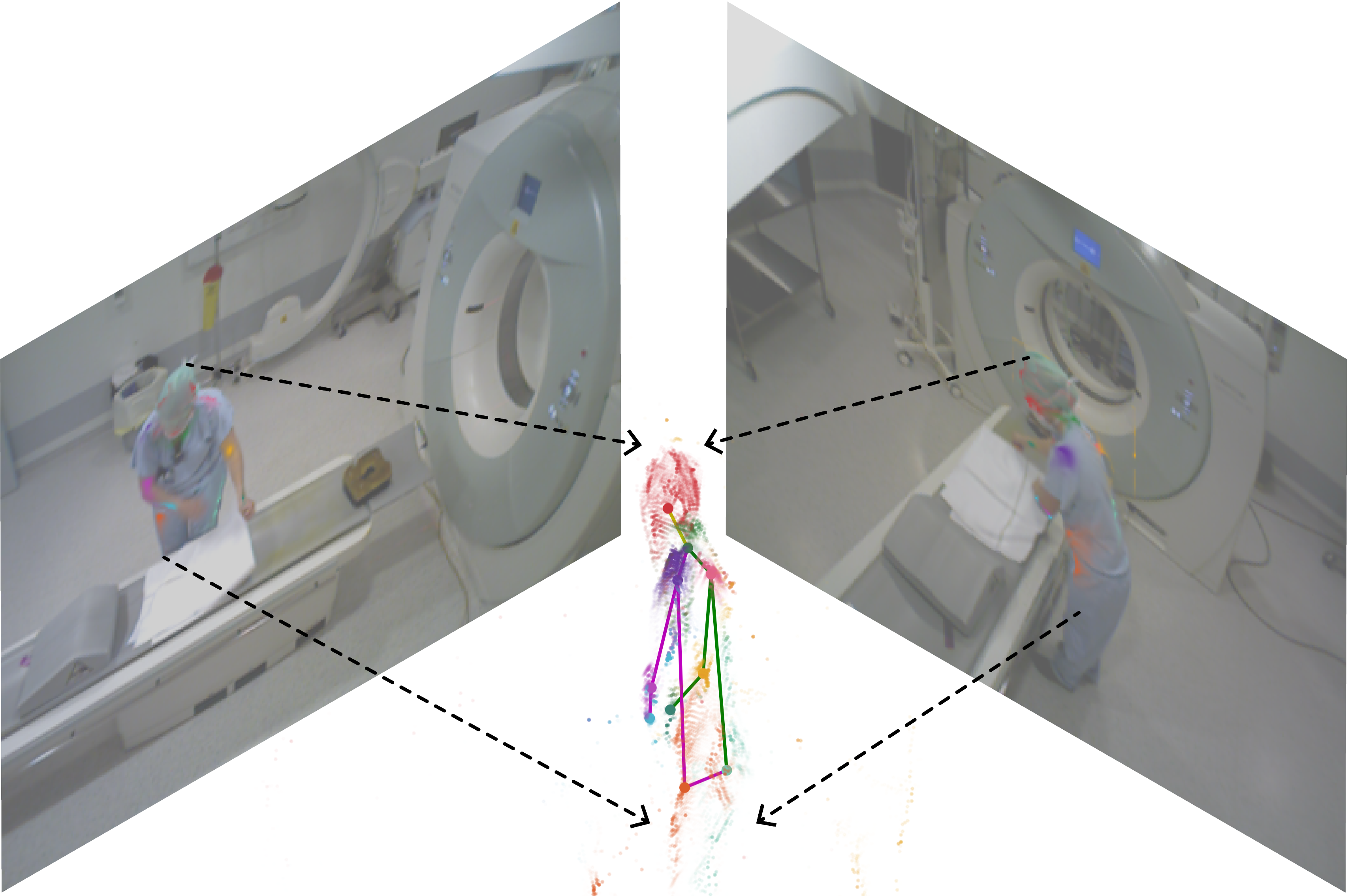}}}
        
    \caption{Previous works find 2D skeletons in each view and project the single body joints to 3D (a). The proposed method projects all heatmap pixels to 3D before skeleton formation (b).}
     \label{fig:projection}
\end{figure}

\begin{figure}
    \includegraphics[width=\textwidth]{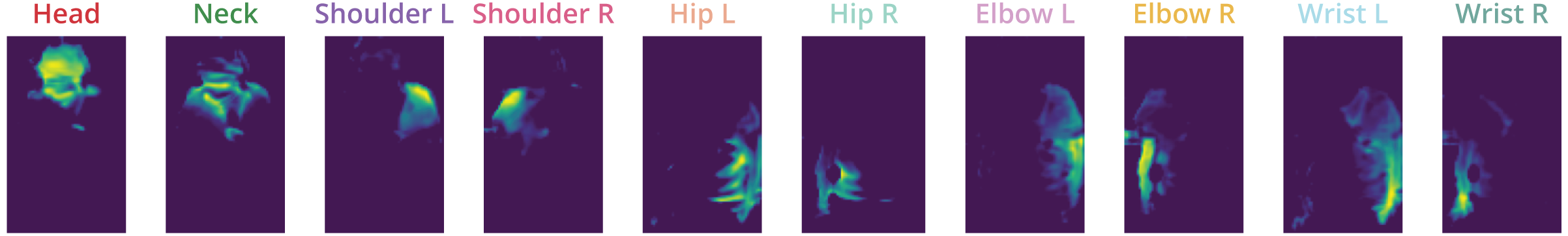}
    \caption{Example of 2D heatmaps that are generated by a U-Net with 10 output channels. We have cut-out the bounding boxes from the full sized heatmaps.}
    \label{fig:channels}
\end{figure}

\subsection{2D and 3D Loss Functions}
Here, we investigate how optimisation based on a 3D loss compares to optimising only in the 2D image domain. To obtain a prediction of a body joint position in a 2D view, we directly aggregate pixel-wise predictions by computing their centre of mass in 2D. To obtain the 3D locations of these 2D positions, we index the corresponding depth images and project the skeletons via camera parameters accordingly. We fuse the 3D skeleton predictions from multiple views by softmax activation over heatmap values indexed with the 2D positions. Figure~\ref{fig:projection} visualises the difference between this approach (a) and the proposed approach, as presented in the previous sections (b).

Table~\ref{tab:losses} shows the results for the approach with a 2D loss and the proposed approach with a 3D loss. We average the MPJPE scores for body joint types that have a left and a right part. The proposed method predicts 3D body joint locations with an average MPJPE of 8.3~cm, with scores ranging from 4.7 to 12.6~cm for heads and wrists, respectively. The upper body (head, neck and shoulders) is recognised with greater precision than the lower body and limbs (hip, elbows, wrists). When optimising in the 2D image domain, the precision decreases to an average MPJPE of 14.3~cm. Particularly for the lower body and limbs, it appears difficult to predict good 3D locations using a 2D loss, with MPJPEs of 16.6, 17.7 and 21.8~cm for hips, elbows and wrists, respectively.

\begin{table}
    \tbl{MPJPE (in cm) for our method optimised in the 2D image domain and, as proposed, directly in 3D using average MPJPE as loss function.}
    {\begin{tabular}{lrrrrrrr} \toprule
        \textbf{Loss} & \textbf{Head} & \textbf{Neck} & \textbf{Shoulder} & \textbf{Hip} & \textbf{Elbow} & \textbf{Wrist} & \textbf{Average} \\
        \midrule
        2D & 10.5 & 8.5 & 11.4 & 16.6 & 17.7 & 21.8 & 14.3 \\
        3D (proposed) & \textbf{5.8} & \textbf{4.7} & \textbf{6.3} & \textbf{10.6} & \textbf{9.7} & \textbf{12.6} & \textbf{8.3} \\
        \bottomrule
    \end{tabular}}
    \label{tab:losses}
\end{table}

Where Figure~\ref{fig:projection}~(b) shows the projected point cloud and predicted 3D skeleton for an unseen sample, Figure~\ref{fig:channels} shows an example of body joint heatmaps generated by one of the U-Nets. Note that the actual heatmaps have the same size as the cropped input images and that we cut out the bounding box for visualisation purposes. Both figures show that the model learns to associate each output channel with a particular body joint type, including a notion of the difference between left and right body parts. Also, the activation maps are visually more similar to segmentation masks than to precise firm activations around the joint locations.

\subsection{Ablation Study}
By ablations, we study the role of the (invertible) data augmentations in the performance of our method with a dataset that has a limited sample size. Table~\ref{tab:augmentation} gives the results when data augmentations are gradually introduced: no augmentation, horizontal flipping, cropping, rotation and colour jitter. The latter involves random changes in brightness, contrast and saturation. It can be seen that all augmentations benefit the results, with an average MPJPE of 10.5~cm for no augmentations to 8.3~cm when all augmentations are applied. Both the geometric augmentations (horizontal flipping, cropping and rotation) and the non-geometric augmentation (colour jitter) improve the results.

The relevance of colour and depth images in the input tensor for finding accurate poses with our method is also evaluated. Table~\ref{tab:modalities} shows the empirical results when leaving out colour or depth images from the input tensor, in comparison with providing both image modalities. Using a single image modality decreases precision to average MPJPEs of 9.8 and 9.3~cm for colour and depth images, respectively. The combination of colour and depth achieves the best results: 8.3~cm. Perhaps surprisingly, using depth images results in better scores than using colour images alone, for all body joint types except wrists. Additionally, the results show that it is possible to use our method solely based on depth images by accepting a precision reduction of 1.0~cm.

\begin{table}
    \tbl{MPJPE (in cm) for our method with and without application of data augmentations.}
    {\begin{tabular}{lrrrrrrr} \toprule
        \textbf{Data augmentation} & \textbf{Head} & \textbf{Neck} & \textbf{Shoulder} & \textbf{Hip} & \textbf{Elbow} & \textbf{Wrist} & \textbf{Average} \\
        \midrule
        No data augmentation    & 7.7 & 5.9 & 7.9 & 12.7 & 12.5 & 16.0 & 10.5 \\
        Horizontal flipping     & 7.1 & 6.1 & 7.7 & 11.8 & 11.5 & 14.3 &  9.7 \\
        Flip + cropping         & 6.5 & 5.5 & 7.1 & 11.7 & 11.1 & 14.2 & 9.4 \\
        Flip + crop + rotation  & 6.7 & 5.4 & 6.7 & 11.3 & 10.0 & 13.0 & 8.8 \\
        Flip + crop + rotat. + colour jitter & \textbf{5.8} & \textbf{4.7} & \textbf{6.3} & \textbf{10.6} & \textbf{9.7} & \textbf{12.6} & \textbf{8.3} \\
        \bottomrule
    \end{tabular}}
    \label{tab:augmentation}
\end{table}

\begin{table}
    \tbl{MPJPE (in cm) for the model with different modalities for input data, i.e. depth and/or colour images.}
    {\begin{tabular}{lrrrrrrr} \toprule
        \textbf{Input} & \textbf{Head} & \textbf{Neck} & \textbf{Shoulder} & \textbf{Hip} & \textbf{Elbow} & \textbf{Wrist} & \textbf{Average} \\
        \midrule
        Colour          & 7.8 & 5.7 & 7.8 & 12.2 & 11.2 & 14.3 & 9.8 \\
        Depth           & 6.1 & 5.4 & 6.9 & 11.1 & 10.9 & 15.1 & 9.3 \\
        Colour + depth  & \textbf{5.8} & \textbf{4.7} & \textbf{6.3} & \textbf{10.6} & \textbf{9.7} & \textbf{12.6} & \textbf{8.3} \\
        \bottomrule
    \end{tabular}}
    \label{tab:modalities}
\end{table}

\subsection{Bounding Boxes from Detection Transformer} \label{sec:detr}
The ablation study was performed with annotated bounding boxes to imitate a data pipeline with an ideal person detector. In reality, however, detectors make mistakes that affect the performance of the pose estimator. We evaluate a joint architecture that consists of a Detection Transformer (DETR) \citep{carion2020end} person detector and the proposed method, trained separately.

We use a standard pre-trained DETR, with a ResNet-101-DC5 backbone, which we fine-tuned on bounding box annotations in the MVOR dataset. For each fold in the cross-validation, we fine-tune a DETR with the training samples of that fold such that all test samples remain unseen to the person detector. The DETR is trained for 10 epochs with a batch size of 2, learning rate 1e-4, backbone learning rate 1e-5, and a weight decay of 1e-4.

To match detected bounding boxes from separate views that belong to the same person, we use a simple algorithm based on 3D locations of the detected boxes. This algorithm projects the centre of all bounding boxes to 3D using camera parameters. Then, it calculates the Euclidean distances between the box centres for all possible combinations of two or three boxes. The whole combination is discarded if the distance between any two boxes in a combination is above threshold $t$. The algorithm selects the combinations with the shortest mean distance, preferring combinations that exist out of three boxes. Unmatched bounding boxes are added to the list as single combinations. For evaluation, we match the annotated boxes with combinations based on the largest intersection over union (IoU), averaged over three camera views. When a combination exists out of one or two boxes, we scored an IoU of 0.0 for the missing boxes. We found $t=75.0$~cm to provide satisfying results by visual inspection.

\begin{table}[b]
    \tbl{MPJPE (in cm) per number of supporting views for a two-step approach \citep{srivastav2018} compared to ours, evaluated on the MVOR dataset. Following the convention, the average MPJPE is calculated over shoulders, hips, elbows and wrists, excluding heads and necks. Standard deviations are calculated over a four fold cross validation.}
    {\begin{tabular}{lrrrrrrrrr}
        \toprule
        \textbf{Joint} & \multicolumn{2}{c}{\textbf{One view}} & \multicolumn{2}{c}{\textbf{Two views}} & \multicolumn{2}{c}{\textbf{Three views}} \\
        \textbf{type} & MV3DReg & Ours & MV3DReg & Ours & MV3DReg & Ours \\
        \midrule
        Shoulder    & 14.4 & \textbf{13.6} ($\pm$ 6.2) & \textbf{ 8.1} &  9.2 ($\pm$ 1.5) &  \textbf{4.9} & 5.6 ($\pm$ 0.7) \\
        Hip         & 29.9 & \textbf{19.6} ($\pm$ 6.4) & 16.0 & \textbf{13.2} ($\pm$ 1.5) &  \textbf{9.9} & 10.1 ($\pm$ 1.5) \\
        Elbow       & 27.3 & \textbf{18.2} ($\pm$ 7.2) & 19.4 & \textbf{15.0} ($\pm$ 2.4) & 10.5 & \textbf{8.8} ($\pm$ 1.4) \\
        Wrist       & 36.1 & \textbf{22.6} ($\pm$ 9.3) & 29.8 & \textbf{17.4} ($\pm$ 0.7) & 14.3 & \textbf{11.9} ($\pm$ 1.4) \\
        \midrule
        Average     & 26.9 & \textbf{18.5} ($\pm$ 6.6) & 18.3 & \textbf{13.7} ($\pm$ 1.1) &  9.9 & \textbf{9.1} ($\pm$ 1.2) \\
        \bottomrule
    \end{tabular}}
    \label{tab:comparison}
\end{table}

Table~\ref{tab:comparison} compares our joint architecture with the MV3DReg method by \citep{srivastav2018}, which is a two-step approach where skeletons are estimated in 2D before these estimations are combined into 3D predictions. We follow the convention from \citep{kadkhodamohammadi2017} by reporting the scores per number of supported views and only for shoulders, hips, elbows and wrists. A view is considered to support a pose whenever the person is visible in that view, i.e. there exists a bounding box from where the pose can be estimated. Similar to the related method, the proposed approach finds better poses when more supporting views are available, with average MPJPEs of 18.5, 13.7 and 9.1~cm for one, two and three views, respectively. Although the performance of our method is slightly better than that of the MV3DReg for poses with three supported views (average MPJPE from 9.9 to 9.1~cm), there is a clear improvement for poses visible in one (26.9 to 18.5~cm) or two (18.3 to 13.7~cm) views. In particular, the estimations of limb locations (elbows and wrists) benefit most from our approach. Standard deviations over the four-fold cross-validation show the largest variations in prediction outcomes of poses supported by one view (standard deviations of 6.2 to 9.3~cm). It appears that the model trained on folds with fewer training data have more problems generalising to unseen poses supported by single views than those supported by multiple views.

\begin{figure}
    \includegraphics[width=\textwidth]{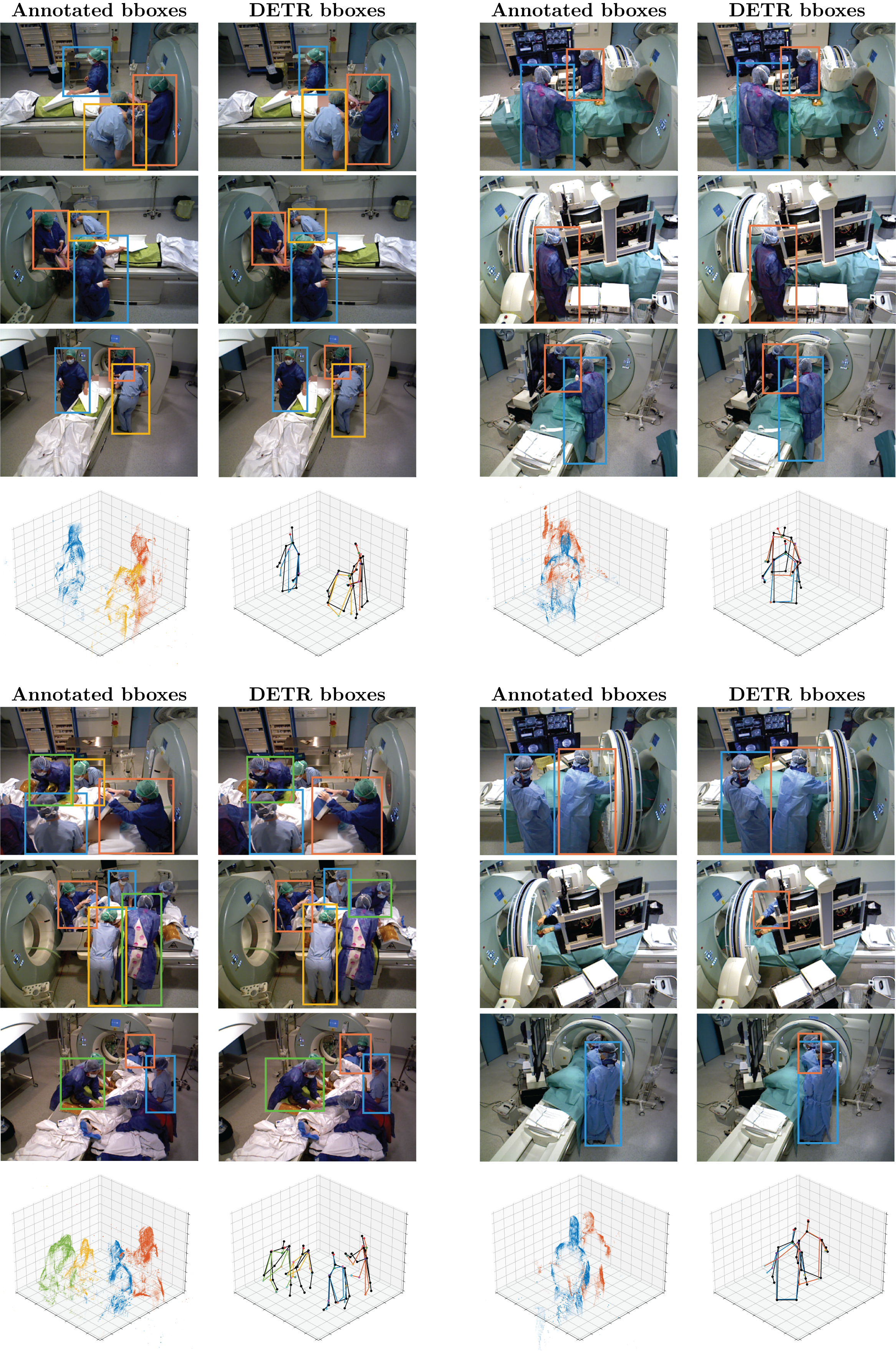}
    \caption{Four example predictions of the proposed method based on DETR person detections. Left column displays annotated bounding boxes, while the right column shows matched DETR bounding boxes. Underneath, 3D plots show the projected point clouds and skeletons (predictions in colour, ground truth in black).} \label{fig:predictions}
\end{figure}

\subsection{Qualitative Analysis} \label{sec:qualitative}
Figure~\ref{fig:predictions} shows four visual examples of predictions by the model as described in Section~\ref{sec:detr}. For each sample, the annotated bounding boxes and the matched detected bounding boxes are given in the left and right columns, respectively. Underneath, two 3D plots display the heatmap point clouds and the estimated poses for multiple persons in the same scene. The upper two examples show the surprisingly good performance of the DETR person detector and bounding box matching, resulting in accurate pose estimations. Here, the method can distinguish people that are close to each other (upper left) and can detect poses for people fully draped in sterile clothing (upper right). The bottom left example displays a crowded scene, resulting in persons that are not detected or have their identities wrongly matched. Although the predicted arm locations are not always accurate, the method remains able to find the right torso locations. The bottom right example shows how DETR is able to find bounding boxes that were not annotated in the MVOR dataset, even when a small part of the body is visible. The 3D plots of this example reveal how our method can generate mistakes: body joints can drift to wrong locations due to noise in the point cloud. This effect can result in anatomically implausible poses, such as the extensive length of the left arm in the estimated pose of the right (orange) person.

\section{Discussion and Conclusion}
In this work, we presented a novel approach for 3D HPE in OR videos, based on end-to-end optimisation of 2D U-Nets with a 3D loss by differentiable projection through camera projection matrices. We found that optimisation directly in 3D is significantly better than optimising in the 2D image domain, as is conventionally done.

Our pose estimation method is completely differentiable and thus end-to-end trainable. This includes projecting 2D heatmaps into point clouds, transforming point clouds into 3D coordinates, but also data augmentation in the multi-view setting, based on inverting the augmentation after the U-Nets are applied. All implemented geometric data augmentations, including horizontal flipping, cropping and rotation, were inverted and resulted in better pose estimations. A non-geometric data augmentations, i.e. colour jitter, can further improve the results. The ability to apply data augmentation is particularly relevant to medical datasets, where the amount of data is generally limited. We hope the approach helps to overcome this limitation for medical multi-view datasets.

We found that depth images are more relevant than colour images for estimating 3D poses with our heatmap-based method. Of all six body joint types, only wrist locations are estimated more accurately with colour images than with depth values. Using depth as single data modality, instead of a multi-modality input, decreases performance with an average MPJPE of 1.0~cm. Since depth images are considered an anonymous data format, in contrast to colour images, it is promising that pose estimation performance decreases only by a small amount when discarding colour imaging entirely. Further research is needed to evaluate the use of depth images in other 3D HPE methods and in other computer vision tasks in the OR, e.g. person detection or tracking.

Fine-tuning a Detection Transformer (DETR) on MVOR data resulted in surprisingly accurate person detections. Bounding boxes from multiple camera views that belong to the same person could easily be combined using a simple algorithm for bounding box matching. As the algorithm is based on depth images, it increases the relevancy of recording depth images in future OR video datasets. We showed that a joint architecture, consisting of DETR and our pose estimator, performs better than a related two-step method \citep{srivastav2018} in 3D HPE in the MVOR dataset. The performance increase is mainly present for poses visible in one or two views, typically involving people occluded by screens, monitor devices or surgical lamps. Since occlusion is one of the main challenges of OR HPE, it is convenient that our method achieves the most significant performance improvements for these poses. The proposed method estimated poses based on detected bounding boxes with average MPJPEs of 18.5, 13.7 and 9.1~cm for poses in one, two and three views, respectively, averaging over shoulders, hips, elbows and wrists.

Our method is purely data-driven and does not include any learned or explicit priors on anatomically plausible skeletons or temporal information to predict the locations of occluded body joints or to smooth out noise in the prediction. Previous work by \cite{hansen2019} has shown that post-processing with a convolutional autoencoder pre-trained on a large external dataset could provide strong anatomical priors, resulting in an error of 8.3~cm, compared to our error of 9.8~cm. We hypothesise that the addition of pose regularisation or post-processing is complementary to our proposed end-to-end differentiable pose estimator, and that including such information in future work could further improve performance.

Qualitative inspection of the results showed that our data pipeline can detect the majority of the people in the OR and match their identities from various camera views. Our method can provide accurate pose estimations in challenging situations, including crowded scenes, heavy occlusions and people in sterile clothing. However, the method is prone to noise around a person's point cloud, resulting in drifting body joints and anatomically implausible poses.

Limitations include the need to calibrate the cameras to find their intrinsic and extrinsic parameters. Ideally, one would use a method that is agnostic to camera locations, such that the method is no longer susceptible to wrongly calibrated camera parameters and to avoid the manual labour of camera calibration. Additionally, it is unlikely that our method will generalise well to other camera positions than the fixed locations with which the method is trained. Datasets that have variation in camera position, but also in surgical procedure and OR type, will help to build algorithms that can generalise well to unseen environments. Last, we trained our method from random initialisation while the dataset is limited in size. Likely, it is beneficial to pre-train the U-Nets on 2D or 3D localisation of human poses with an external dataset.

With the proposed elements, we see the potential for accurate 3D pose estimators that are specifically designed for the complex surgical environment. In particular, end-to-end methods that optimise directly in 3D using differentiable camera projections are a promising direction for future methods. Consequently, we envision that OR pose estimators will be used to build higher-level algorithms that make the OR safer, more efficient and more pleasant.











\bibliographystyle{tfcse}
\bibliography{main}

\end{document}